\documentclass[copyright,noncommercial]{eptcs}
\pdfoutput=1
\providecommand{\volume}{5}
\providecommand{\anno}{2009}
\providecommand{\firstpage}{97}
\providecommand{\eid}{8}
\providecommand{\event}{Y. Deville and C. Solnon (Eds): Sixth International Workshop\\
 on Local Search Techniques in Constraint Satisfaction (LSCS'09).}

\usepackage{ifpdf}
\usepackage{hevea}

\newif\iflatex
\latexfalse

\ifhevea 
      \pdffalse 
\else 
      \ifpdf 
      \else
      \latextrue  
      \fi
\fi

\usepackage[pdftex]{graphicx}
\usepackage{hyperref}

\newcommand{\InsertImage}[2][1]%
   {\centerline{\fbox{{\includegraphics[width=#1\columnwidth]{#2}}}}}

\newcommand{\InsertImageNoBox}[2][1]%
   {\centerline{\includegraphics[width=#1\columnwidth]{#2}}}

\ifpdf

\newcommand{\MyEMail}[2]{#2\footnote{\texttt{#1}}}
\else
\ifhevea

\newcommand{\MyEMail}[2]{\@aelement{HREF="mailto:#1"}{#2}}
\else

\fi
\fi

\usepackage{amsmath}

\def\PROG{\textsf{AS/Cell}}

\def\CF#1{\multicolumn{1}{||c|}{#1}}
\def\CFD#1{\multicolumn{1}{||c||}{#1}}
\def\C#1{\multicolumn{1}{c|}{#1}}
\def\D#1{\multicolumn{1}{c||}{#1}}

\newcommand{\AddImage}[4]%
{\begin{figure}[htb]%
\InsertImageNoBox[#3]{#2}%
\caption{#1}\label{#4}\end{figure}}

\newcommand{\AddGraph}[3]%
{\AddImage{#1}{graphs/#2}{#3}{#2-graph}}

\newcommand{\AddProblemOverallTable}[2]%
{\begin{table}[htbp]
\begin{center}\begin{tabular}{||c|r|r|r|r|r|r|r||}%
\hline%
\CF{size} & \C{time}  & \multicolumn{5}{c|}{speedup with $k$ SPUs} & \D{time}     \\
\cline{3-7}
\CF{}     & \C{1 SPU} & \C{2}  & \C{4}  & \C{8}  & \C{12} & \C{16} & \multicolumn{1}{@{~}c@{~}||}{16 SPUs} \\
\hline\hline
\input{table-data/#2-avg.tex}
\hline%
\end{tabular}\end{center}\caption{timings (sec) and speedups for #1}\label{#2-table}\end{table}
}

\newcommand{\AddProblemOverallTableAndGraph}[2]%
{\begin{table}[htbp]
\begin{center}\begin{tabular}{||c|r|r|r|r|r|r|r||}%
\hline%
\CF{size} & \C{time}  & \multicolumn{5}{c|}{speedup with $k$ SPUs} & \D{time}     \\
\cline{3-7}
\CF{}     & \C{1 SPU} & \C{2}  & \C{4}  & \C{8}  & \C{12} & \C{16} & \multicolumn{1}{@{~}c@{~}||}{16 SPUs} \\
\hline\hline
\input{table-data/#2-avg.tex}
\hline%
\end{tabular}\end{center}\caption{timings (sec) and speedups for #1}\label{#2-table}\end{table}

\AddGraph{average time with 1 and 16 SPUs for #1}{#2-scaling-avg}{0.95}
}

\newcommand{\AddProblemInstanceDetailTableAndGraph}[2]%
{\begin{table}[htbp]
\begin{center}\begin{tabular}{||c||r|r||r|r||}%
\hline%
\CFD{\#SPUs} & \multicolumn{2}{c||}{average case} & \multicolumn{2}{c||}{worst case} \\%
\cline{2-3} \cline{4-5}
\CFD{}       &  \C{time (sec)}   &  \D{speedup}   &  \C{time (sec)}  &  \D{speedup}  \\
\hline\hline%
\input{table-data/#2.tex}%
\hline%
\end{tabular}\end{center}\caption{average and worst times for #1}\label{#2-table}\end{table}

\AddGraph{average and worst timings for #1}{#2-speedup}{1.2}
}

\newcommand{\AddProblemInstanceDetailTable}[2]%
{\begin{table}[htbp]
\begin{center}\begin{tabular}{||c||r|r||r|r||}%
\hline%
\CFD{\#SPUs} & \multicolumn{2}{c||}{average case} & \multicolumn{2}{c||}{worst case} \\%
\cline{2-3} \cline{4-5}
\CFD{}       &  \C{time (sec)}   &  \D{speedup}   &  \C{time (sec)}  &  \D{speedup}  \\
\hline\hline%
\input{table-data/#2.tex}%
\hline%
\end{tabular}\end{center}\caption{average and worst times for #1}\label{#2-table}\end{table}

}

\title{Parallel local search for solving Constraint Problems on the Cell Broadband Engine (Preliminary Results)}

\author{Salvador Abreu
\institute{Universidade de \'Evora and CENTRIA FCT/UNL\\
  Portugal}
  \email{spa@di.uevora.pt}
  \and
  Daniel Diaz
  \institute{University of Paris 1-Sorbonne\\
  France}
  \email{Daniel.Diaz@univ-paris1.fr}
  \and
  Philippe Codognet
  \institute{JFLI, CNRS / UPMC / University of Tokyo, \\
  Japan}
  \email{Philippe.Codognet@lip6.fr}
  }
\def\titlerunning{Local Search on Cell}
\def\authorrunning{Abreu, Diaz \& Codognet}

\begin{document}

\maketitle

\begin{abstract}
  We explore the use of the Cell Broadband Engine (Cell/BE for short)
  for combinatorial optimization applications: we present a parallel
  version of a constraint-based local search algorithm that has been
  implemented on a multiprocessor BladeCenter machine with twin
  Cell/BE processors (total of 16 SPUs per blade).  This algorithm was
  chosen because it fits very well the Cell/BE architecture and
  requires neither shared memory nor communication between processors,
  while retaining a compact memory footprint.  We study the
  performance on several large optimization benchmarks and show that
  this achieves mostly linear time speedups, even sometimes
  super-linear.  This is possible because the parallel implementation
  might explore simultaneously different parts of the search space and
  therefore converge faster towards the best sub-space and thus
  towards a solution.  Besides getting speedups, the resulting times
  exhibit a much smaller variance, which benefits applications where a
  timely reply is critical.
\end{abstract}

\section{Introduction}
\label{sec:introduction}

The Cell processor has shown its power for graphic and server
applications, and more recently has been considered as a good
candidate for scientific calculations~\cite{cell}.  Its floating point
arithmetic performance and energy efficiency make it useful as a basic
block for building super-computers, cf.~the ``Roadrunner'' machine
based on Cell processors which is currently the fastest supercomputer.
However, its ability to perform well for general-purpose applications
has been questioned, and Cell programming has always been considered
as very challenging.  We investigate in this paper the use of the
Cell/BE for combinatorial optimization applications and
constraint-based problem solving.  It is worth noticing that in these
domains most of the attempts to take advantage of the parallelism
available in modern multi-core architectures have targeted homogeneous
systems, for instance Intel or AMD-based machines and make use of
shared memory,
e.g.~\cite{DBLP:conf/cp/Perron99,DBLP:journals/tse/HolzmannB07,manySAT}.
The different cores are working on shared data-structures which
somehow represent a global environment in which the subcomputations
are taking place.  Such an approach cannot be used for Cell-based
machines, because heavy use of shared memory would degrade the overall
performance of this particular multi-core system: in order to extend
the use of the Cell processor for combinatorial optimization and
constraint-based problem solving, new approaches have to be
investigated, in particular those that can lead to independent
subcomputations requiring little or no communication between
processing units and limited or even no accesses to the main (shared)
memory.  We decided to focus on Local Search algorithms, also called
``metaheuristics'', which have attracted much attention over the last
decade from both the Operations Research and the Artificial
Intelligence communities, in order to tackle very large combinatorial
problems which are out of range for the classical exhaustive search
methods.  Local search and metaheuristics have been used in
Combinatorial Optimization for finding optimal or near-optimal
solutions and have been applied to many different types of problems
such as resource allocation, scheduling, packing, layout design,
frequency allocation, etc.

To enable the use of the Cell/BE for combinatorial optimization
applications, we have developed a parallel extension of a
constraint-based local search algorithm based on a method called
``Adaptive Search'' which was proposed a few years ago
in~\cite{DBLP:conf/saga/CodognetD01,mic/CodognetD03}.

To assess the viability of this approach, we experimented on several
classical benchmarks of the constraint programming community from
CSPlib~\cite{CSPLIB}.  These structured problems are somehow
abstractions of real problems and are therefore representative of
real-life applications; they are classically used in the community for
benchmarking new methods.  The preliminary implementation results for
the parallel Adaptive Search method show a good behavior when scaling
up the number of cores (from one to sixteen): speedups are, most of
`the time, practically linear, especially for large-scale problems and
our experiments even exhibit a few super-linear speedups because the
simultaneous exploration of different subparts of the search space
may converge faster towards a solution.

Another interesting point to mention is that all experiments show a
better robustness 
of the results on the multi-core version when compared to the
sequential algorithm, as will be explained below.  Because local
search methods make use of randomness for the diversification of the
search, execution times may vary from one run to another.  This is
why, when benchmarking such methods, execution times have to be
averaged on many runs (in our experiments, we always take the average
of 50 runs).  Our implementation results show that for a parallel
version running on 16 cores, the difference between the minimal and
maximal execution times, as well as the overall variance of the
results, decreases significantly with respect to the reference
sequential implementation.  The main result of this is that execution
times become more predictable and this is, of course, an advantage in
real-time applications with bounded response time requirements.

The remainder of this article is organized as follows: after an
introduction, section~\ref{sec:adapt-search-algor} discusses the
Adaptive Search algorithm and its parallel version is presented in
section~\ref{sec:as-par-algo}.  We proceed with a performance analysis
in section~\ref{sec:simple-perf-eval}, which is analyzed and commented
on in section~\ref{sec:analys-perf-robustn}.  Finally, we conclude in
section~\ref{sec:conclusion} and present our lines for related future
research.

\section{Parallelizing Constraint Solvers}
\label{sec:parallelization}

Parallel implementation of search algorithms has a long history,
especially in the domain of logic programming, see~\cite{DBLP:journals/csur/KergommeauxC94}
for an overview.  Most of the proposed implementations are based on
the so-called OR-parallelism, splitting the search space between
different processors but making used of a shared or duplicated stack
for coping with the adequate execution environment and they rely on a
Shared Memory Multiprocessor (SMP) architecture for the parallel
execution support.  For some years, similar techniques have been used
for Model Checkers, which are used as verification tools for hardware
and software, such as the SPIN software~\cite{LTL,DBLP:journals/tse/HolzmannB07}.  These
implementations are also based on some kind of OR-parallelism and,
again, these approaches are well-suited for multi-core architecture
with shared memory.  More recently, there have been several
initiatives to extend SAT solvers for parallel machines, in particular
multi-cores \cite{manySAT,PminiSAT,PaMIraXT}.  However these
frameworks require a shared memory model and will thus not be scalable
to massively parallel machines or architectures for which
communication through (distributed) shared memory is costly such as a
cluster system or for hetherogeneous multicore processors such as the
Cell/BE.  It is worth noticing that now some authors are also
extending SAT solvers to PC cluster architectures~\cite{c-sat}, using
a hierarchical shared memory and trying to minimize communication
between clusters.
While moving from traditional SMP machines to multi-core systems is a
relatively straightforward change, it is not necessarily so for more
exotic architectures, such as the hetherogeneous multicore chips which
include the Cell/BE.

For Constraint Satisfaction Problems, early work has been done in the
context of Distributed Artificial Intelligence and multi-agent
systems, see for instance~\cite{DBLP:journals/tkde/YokooDIK98}, but
these methods, even if interesting from a theoretical point of view,
cannot lead to efficient algorithms and cannot compete with good
sequential implementations.  Moreover, the focus is usually not on
performance but on the formulating a problem in a distributed fashion.
Only very few implementations of efficient constraint solvers on
parallel machines have been reported, most
notably~\cite{DBLP:conf/cp/Perron99}, which again is aimed at
shared-memory architectures and recently ~\cite{comet06} which
proposes a distribted extendion of the Comet local search solver for
clusters of PCs.

\section{The Adaptive Search Algorithm}
\label{sec:adapt-search-algor}


Over the last decade, the application of local search techniques for
constraint solving in general (and not only for combinatorial
optimization) has started to draw some interest in the CSP community.
A generic, domain-independent local search method named Adaptive
Search was proposed by~\cite{DBLP:conf/saga/CodognetD01,mic/CodognetD03}, a new meta-heuristic that
takes advantage of the structure of the problem in terms of
constraints and variables and can guide the search more precisely than
a global cost function to optimize (such as for instance the number of
violated constraints).  The algorithm also uses an short-term adaptive
memory in the spirit of Tabu Search in order to prevent stagnation in
local minima and loops.  This method is generic, can be applied to a
large class of constraints (e.g. linear and non-linear arithmetic
constraints, symbolic constraints, etc) and naturally copes with
over-constrained problems.  The input of the method is a problem in
CSP format, that is, a set of variables with their (finite) domains of
possible values and a set of constraints over these variables.  For
each constraint, an ``error function'' needs to be defined; it will
give, for each tuple of variable values, an indication of how much the
constraint is violated.  For instance, the error function associated
with an arithmetic constraint $|X - Y| < c$, for a given constant $c
\geq 0$, can be $max(0, |X-Y|-c)$.  Adaptive Search relies on
iterative repair, based on variable and constraint error information,
seeking to reduce the error on the worst variable so far.  The basic
idea is to compute the error function for each constraint, then
combine for each variable the errors of all constraints in which it
appears, thereby projecting constraint errors onto the relevant
variables.  Finally, the variable with the highest error will be
designated as the ``culprit'' and its value will be modified.  In this
second step, the well known min-conflict heuristic~\cite{min-conflict}
is used to select the value in the variable domain which is the most
promising, that is, the value for which the total error in the next
configuration is minimal.  In order to prevent being trapped in local
minima, the Adaptive Search method also includes a short-term memory
mechanism to store configurations to avoid (variables can be marked
Tabu and "frozen" for a number of iterations), and also integrates
restart-based transitions to escape stagnation around local minima.
Restarts are partial and are guided by the number of variables being
marked Tabu.  The core ideas of adaptive search can be summarized as
follows:
\begin{itemize}

\item to consider for each constraint a heuristic function that is
  able to compute an approximated degree of satisfaction of the goals
  (the current ``error'' on the constraint);

\item to aggregate constraints on each variable and project the error
  on variables thus trying to repair the ``worst'' variable with the most
  promising value;

\item to keep a short-term memory of bad configurations to avoid
  looping (i.e. some sort of ``tabu list'').

\end{itemize}

\subsection*{Algorithm}

Consider an n-ary constraint $c(X_1, \cdots X_n)$ and associated variable
domains $D_1, \cdots D_n$.  An error function $f_c$ associated to the
constraint $c$ is a real-valued function from $D_1 \times \cdots \times D_n$
such that $f_c(X_1, \cdots X_n)$ has value zero if $c(X_1, \cdots X_n)$ is
satisfied.  The error function will in fact be used as a heuristic value to
represent the degree of satisfaction of a constraint and will thus give an
indication on how much the constraint is violated.  This is very similar to
the notion of ``penalty functions'' used in continuous global optimization.
This error function can be seen as (an approximation of) the distance of the
current configuration to the closest satisfiable region of the constraint
domain.  Since the error is only used to heuristically guide the
search, we can use any approximation when the exact distance is difficult (or
even impossible) to compute.\\
\ 
\\
\underline{Input}\\
\\
Problem given in CSP format:\\
-   a set of variables $V=\{V_1, V_2, ... , V_n\}$ with associated domains of values\\
-   a set of constraints $C=\{C_1, C_2, ... , C_k\}$ with associated error functions\\
-   a combination function to project constraint errors on variables\\
-   a (positive) cost function to minimize\\
\\
Some tuning parameters:\\
-   T : Tabu tenure (number of iterations a variable is frozen)\\
-   RL : reset limit (number of frozen variables triggering a reset)\\
-   RP : reset percentage (percentage of variables to reset)\\
-   Max\_I : maximal number of iterations before restart\\
-   Max\_R : maximal number of restarts\\
\\
\underline{Output}\\
\ \\
A solution (configuration where all constraints are satisfied) if the CSP is satisfied or to a quasi-solution of minimal cost otherwise.\\
\\
\underline{Algorithm}

\begin{tabbing}
xx \= xx \= xx \= xxxx \= xxxx \= \kill
Iteration = 0\\
Restart = 0\\
{\bf Repeat}\\
\> Restart = Restart + 1\\
\> Iteration = Iteration +1\\
\> Tabu\_Nb = 0\\
\> Compute a random assignment A of variables in V\\
\> Opt\_Sol = A\\
\> Opt\_Cost = cost(A)\\
\> {\bf Repeat}\\
\> \> 1. \> Compute errors of all constraints in C \\
\> \> \>    and combine errors on each variable\\
\> \> \> (by considering only the constraints in which a variable appears)\\
\> \> 2. \> select the variable X (not marked Tabu) with highest error\\
\> \> 3. \> evaluate costs of possible moves from X\\
\> \> 4. \> {\bf if} \> no improvment move exists\\
\> \> \>    {\bf then} \> mark X as Tabu until iteration number: Iteration + T\\
\> \> \> \>               Tabu\_Nb = Tabu\_Nb + 1\\
\> \> \> \>               {\bf if} \> Tabu\_Nb $>$ RL\\
\> \> \> \>               {\bf then} \> randomly reset RP variables in V \\
\> \> \> \>                          \> (and unmark those which are Tabu)\\
\> \> \>   {\bf else} \> select the best move and change the value of X\\
\> \> \> \>              accordingly to produce next configuration A'\\
\> \> \> \>              {\bf if} \> cost(A')  $<$ Opt\_Cost\\
\> \> \> \>              {\bf then} \> Opt\_Sol = A'\\
\> \> \> \> \>                         Opt\_Cost = cost(A')\\
\> {\bf until} a solution is found or Iteration  $>$ Max\_I\\
{\bf until} a solution is found or Restart $>$  Max\_R\\
output (Opt\_Sol, Opt\_Cost)
\end{tabbing}

Adaptive Search is a simple algorithm but it turns out to be quite
efficient in practice~\cite{mic/CodognetD03}.  Considering the
complexity/efficiency ratio, it can be a very effective way to
implement constraint solving techniques in larger software tools,
especially for anytime algorithms where (approximate) solutions have
to be computed within a limited amount of time.

\section{Parallel Algorithm on the Cell/BE}
\label{sec:as-par-algo}

We will not present the Cell/BE processor architecture here, some
features of this architecture, however, deserve mention because they
strongly shape what applications may succeed when ported:
\begin{itemize}
\item A hybrid multicore architecture, with a general-purpose
  ``controller'' processor (the PPE, a PowerPC instance) and eight
  specialized processors (SPEs.)
\item Two Cell/BE processor chips may be linked to appear as a
  multiprocessor with 16 SPEs.
\item The PPEs are connected via a very high-bandwidth internal bus,
  the EIB.
\item The PPEs may only perform operations on their local store, which
  contains both code and data and is limited to 256KB.
\item The PPEs may access system memory and each other's private
  memory by means of DMA operations.
\end{itemize}
The interested reader can refer to the IBM Redbook~\cite{cell} for
further information on this architecture, as well as the performance
and capacity tradeoffs which affect Cell/BE programs.



The basic idea in extending the algorithm for parallel implementation
is to have several distinct parallel search engines for exploring
simultaneously different parts of the search space, and to start each
such engine a a different processor.  This is very natural to achieve
with the Adaptive Search algorithm: one just needs to start each
engine with a different, randomly computed, initial configuration,
that is, a different assignment of values to variables.  Subsequently,
each ``Adaptive Search engine'' can perform the sequential algorithm
described in the previous section independently.  As soon as one
process finds a solution, or when all processors reach the maximal
number of iterations allowed, all processors are halted and the
algorithm finishes with the condition that led to the termination
(solution found or maximum iterations reached in all processors).

\subsection*{The Parallel Algorithm}

The Cell/BE processor architecture is reflected on the task structure,
in which a controller thread resides in the PPE and each SPE has a
worker thread.
\begin{itemize}
\item The PPE gets the real time $T0$, launches a given number of
  threads, each with an identical SPU context, and then waits for a
  solution.
\item Each SPE starts with a random configuration (held entirely in
  its local storage) and improves it step by step, applying the
  algorithm of section~\ref{sec:adapt-search-algor}.
\item As soon as an SPE finds a solution, it sends it to the main
  memory (using a DMA operation) and informs the PPE.
\item The PPE then propagates this information to all other SPEs to
  stop their job and waits until all SPUs have finished (join).  After
  that, it gets the real time $T1$ and provides both the solution and
  the execution time~\footnote{the execution time is then the real
    elapsed time since the beginning of the program until the join
    (thus including SPUs initialization and termination phases).} $T =
  T1 - T0$.
\end{itemize}


\noindent It is worth pointing out that SPEs do not communicate among
themselves and only do so with the PPE upon termination: each SPE can
work blindly on its configuration until it reaches an outcome
(solution or failure).  Indeed, we managed to fit both the program and
the data in the 256KB of local store of each SPU, even for admittedly
large benchmarks.  This turns out to be possible for two reasons: (1)
the simplicity and compactness of the algorithm, (2) the compactness
of the encoding of the combinatorial problem as a CSP, that is,
variables with finite domains and many predefined constraints,
including arithmetics.  This is especially true when compared, for
instance, to a SAT encoding where only boolean variables can be used
and each constraint has to be explicitly decomposed into a set of
boolean formulas yielding problem formulations which easily reach
several thousands of literals.

To summarize, the Adaptive Search method requirements are a good match
for the Cell/BE architecture: \emph{not much data but a lot of
  computation}.

\section{Performance Evaluation}
\label{sec:simple-perf-eval}

We now present and discuss the performance of our implementation of
\PROG{}.  The code running on each SPU is derived from the code used
in \cite{DBLP:conf/saga/CodognetD01,mic/CodognetD03} which is an
implementation of the Adaptive Search for permutation problems.  It is
worth noting that no code specialization has been made to benefit from
the full potential of the Cell processor (namely vectorization, branch
removing, ...)  It is reasonable to expect a significant speedup when
these aspects are taken into account.

Since the Adaptive Search uses random configurations and progression,
each benchmark has been executed 50 times.  There are two interesting
ways for aggregating those results: considering the average case
(average of 50 executions times after removing both the lowest and the
highest times) and considering the worst case (maximum of the 50
executions).  On one hand, the former is classical and gives a precise
idea of the behavior of the \PROG{}.  On the other hand, the latter is
also interesting for real-time applications since it represents the
``worst-case'' one can encounter (too high a value can even prevent
the use in time-critical applications).  Interestingly, \PROG{}
improves both cases, achieving linear speedups and sometimes even
super-linear speedups.


\subsection{The All-Interval series}
\label{sec:all-interval-series}

Although looking like a pure combinatorial search problem, this
benchmark is in fact a well-known exercise in music
composition~\cite{DBLP:journals/soco/TruchetC04}.  The idea is to
compose a sequence of N notes such that all are different and tonal
intervals between consecutive notes are also distinct (see
Figure~\ref{all-interval-ex}).

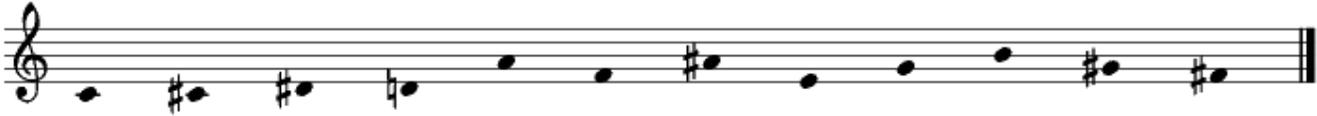
\AddImage{an example of all-interval in music}{all-interval-music}{1.15}{all-interval-ex}

This problem is described as \texttt{prob007} in the
CSPLib~\cite{CSPLIB}.  It is equivalent to finding a permutation of
the N first integers such that the absolute difference between two
consecutive pairs of numbers are all different.  This amounts to
finding a permutation $(X1, \ldots X_N)$ of $\{0, \ldots N-1\}$ such
that the list $(abs(X_1-X_2), abs(X_2-X_3) \ldots abs(X_{N-1}- x_N))$
is a permutation of ${1,\ldots,N-1}$.

Table~\ref{all-interval-table} presents the average time of 50 running
(in seconds) for several instances of this benchmark together with the
speedup obtained when using different number of SPUs.  From this data
one can conclude the speedup linearly increases with the number of
SPUs to reach 11 with 16 SPUs.  This factor appears to be constant
whatever the size of problem.

\AddProblemOverallTable{all-interval series}{all-interval}

It is worth noticing that state-of-art constraint solvers
(e.g.~Gecode) are able to find the trivial solution $(0, N-1, 1, N-2,
2, N-3, \ldots)$ in a reasonable amount of time but fail to find an
interesting solution for $N \ge 20$.  The \PROG{} implementation is
able to find solutions for $N = 450$ in 1.5 minute with 16 SPUs.
Table~\ref{all-interval-450-table} detail this instance
providing information on both the average case and the worst case
(together with the associated speedups).  In this problem, linear
speedups are obtained: with 16 SPUs the average time is 11 times
faster.  When discussing worst cases, the time is divided by a factor
26.

\AddProblemInstanceDetailTable{all-interval 450}{all-interval-450}


\subsection{Number partitioning}
\label{sec:number-partitioning}

This problem consists in finding a partition of numbers $\{1, \ldots N\}$ into
two groups $A$ and $B$ such that:
\begin{itemize}
  \item $A$ and $B$ have the same cardinality
  \item the sum of numbers in $A$ is equal to the sum of numbers in
    $B$
  \item the sum of squares of numbers in $A$ is equal to the sum of
    squares of numbers in $B$
\end{itemize}

\noindent A solution for $N = 8$ is $A = (1,4,6,7)$ and $B =
(2,3,5,8)$ since:

\begin{center}
  $1 + 4 + 6 + 7 = 18 =  2 + 3 + 5 + 8$\\
  $1^2 + 4^2 + 6^2 + 7^2 = 102 = 2^2 + 3^2 + 5^2 + 8^2$ \\
\end{center}

\noindent This problem admits a solution iff $N$ is a multiple of 8
and is modeled with $N$ variables $V_i \in \{1 \ldots N\}$ which form
a permutation of $\{1 \ldots N\}$.  The first $N/2$ variables form the
group $A$, the $N/2$ last variables the group $B$.  There are two
constraints:

\begin{center}
  $\Sigma_{i=1}^{N/2} V_i~=~ N(N+1)/4 ~=~ \Sigma_{i=N/2+1}^{N} V_i$ \\
  $\Sigma_{i=1}^{N/2} V_i^2 ~=~ N(N+1)(2N+1)/12 ~=~ \Sigma_{i=N/2+1}^{N} V_i^2$
\end{center}

The possible moves from one configuration consist in all possible
swaps exchanging one value in the first subset with another one in the
second subset.  The errors on the 2 equality constraints are computed
as the absolute value of the difference between the actual sum and the
expected constant (e.g.~$N(N+1)/4$).  In this problem, like for the
all-intervals example, all variables play the same role and there is
no need to project errors on variables.  The total cost of a
configuration is the sum of the absolute values of both constraint
errors.  A solution is found when the total cost is zero.

Table~\ref{partit-table} details the average running times (in
seconds) for several instances of this problem together with the
speedup obtained when using different numbers of SPUs.  Similarly to
what occured with the all-interval series, the speedup increases
linearly up to a factor of 11.  Again, the speedup appears to be
independent from the size of the problem.

\AddProblemOverallTable{number partitioning}{partit}

Once more, it is worth noticing that Constraint Programming systems
such as GNU Prolog cannot solve this problem for instances larger than
128.  On the other hand the \PROG{} implementation is able to find
solutions for $N = 2600$ in few seconds with 16 SPUs: this problem
scales very well and it is possible to solve even larger instances.
Table~\ref{partit-2600-table} details the largest instance both for
the average case and the worst case (together with the associated
speedups).  For this example the speedups are linear: with 16 SPUs the
average time is divided by 11 while the worst case is divided by 17.

\AddProblemInstanceDetailTable{partit 2600}{partit-2600}


\subsection{The Perfect-Square placement problem}
\label{sec:perf-square-plac}

This problem is described as \texttt{prob009} in CSPLib~\cite{CSPLIB}.
It is also called the \emph{squared square} problem~\cite{Lint92} and
consists in packing a set of squares into a master square in such a
way that no squares overlap each other.  All squares have different
sizes and they fully cover the master square (there is no spare
capacity).  The smallest solution involves 21 squares which must be
packed into a master square of size 112.

Since the system we are basing our work on (Adaptive Search) only
deals with permutation problems, we have modeled this problem as a set
of N variables whose values corresponds to the sizes of the squares to
be placed, in order -- this is not the best modeling but complies with
the requirements of the available implementation.  Each square in a
configuration is placed in the lowest and leftmost possible slot.

Moving from a configuration to another consists in swapping 2
variables.  To compute the cost of a configuration, the squares are
packed as explained above. As soon as a square does not fit in the
lowest/leftmost slot the placement stops.  The cost of the
configuration is a formula depending on several criteria on the set of
non placed squares (number of non-placed squares and the size of the
biggest) and on remaining slots in the master square (sum of heights,
largest height, sum of widths).  As usual, a configuration is a
solution when its cost drops to zero.

We tried 5 different instances of this problem taken
from~\cite{CSPLIB,Bouwkamp92} whose input data are summarized in
table~\ref{tab:perf-sq-instances}.
\begin{table}[htb]
  \centering
  \begin{tabular}{|c|c|c|c|}
    \hline
    problem  & master square & \multicolumn{2}{c|}{squares to place} \\
    \cline{3-4}
    instance & size          & number & largest \\
    \hline
    1 & 112 $\times$ 112 & 21 &  50 $\times$  50 \\
    2 & 228 $\times$ 228 & 23 &  99 $\times$  99 \\
    3 & 326 $\times$ 326 & 24 & 142 $\times$ 142 \\
    4 & 479 $\times$ 479 & 24 & 175 $\times$ 175 \\
    5 & 524 $\times$ 524 & 25 & 220 $\times$ 220 \\
    \hline
  \end{tabular}
  \caption{perfect-square instances}
  \label{tab:perf-sq-instances}
\end{table}
Table~\ref{perfect-square-table} presents the data associated to the
average case for these instances.  Running 16 SPUs, the speedup ranges
from 11 to 16 depending on the instance.

\AddProblemOverallTable{perfect square}{perfect-square}

As previously explained, our modeling is not the best one: a modeling
explicitely using variables to encode X and Y coordinates of each
square would be clearly better as done in a Constraint Programming
modeling.  Nevertheless, \PROG{} performs rather well and the most
difficult instance (number 5) is solved in less than 10 seconds with
16 SPUs.  Table~\ref{perfect-square-5-table} provides more information
for this instance both for the average case and the worst case
(together with the associated speedups).  In this problem linear
speedups are obtained: with 16 SPUs, both the average and worst case
times are about 16 times lower.

\AddProblemInstanceDetailTable{perfect square \#5}{perfect-square-5}


\subsection{Magic squares}
\label{sec:magic-squares}

The magic square problem is listed as \texttt{prob019} in
CSPLib~\cite{CSPLIB} and consists in placing the numbers $\{1,2 \cdots
N^2\}$ on an $N\times N$ square, such that the sum of the numbers in
all rows, columns and the two diagonal is the same.  The constant
value that should be the sum of all rows, columns and the two
diagonals can be easily computed to be $N(N^2+1)/2$.  %

The modeling for \PROG{} involves $N^2$ variables
$X_1,\ldots,X_{N^2}$.  The error function of an equation $X_1 + X_2 +
\ldots + X_k = b$ is defined as the value of $X_1 + X_2 + \ldots + X_k
- b$.  The combination operation is the sum of the absolute values of
the errors.  The overall cost function is the addition of absolute
values of the errors of all constraints.  A configuration with zero
cost is a solution.

Table~\ref{magic-square-table} details the average running times (in
seconds) for several instances of this problem together with the
speedup obtained when using different numbers of SPUs.  Using 16 SPUs,
the obtained speedup increases with the size of the problem to reach
22 for the largest instance.

\AddProblemOverallTableAndGraph{magic squares}{magic-square}

It is worth noticing that this benchmarks is one of the most
challenging: Constraint programming systems such as GNU-Prolog or ILOG
Solver perform poorly on this benchmark and cannot solve instances
greater than $10\times 10$.  On the other hand \PROG{} is able to
solve $100 \times 100$ in only few seconds with 16 SPUs.
Table~\ref{magic-square-100-table} details the largest instance both
for the average case and the worst case with associated speedups.  For
this example the speedups are super-linear: with 16 SPUs the average
time is divided by 22 while the worst case is divided by 500!

\AddProblemInstanceDetailTable{magic squares $100 \times 100$}{magic-square-100}

\section{Analysis: Performance and Robustness}
\label{sec:analys-perf-robustn}

The performance evaluation of section~\ref{sec:simple-perf-eval} has
shown that the Adaptive Search method is a good match for the Cell/BE
architecture.  This processor is clearly a serious candidate to
effectively solve highly combinatorial problems.  All problems tested
were accelerated when using several SPUs.  For 3 of the problems the
ultimate speedup obtained with 16 SPUs seems constant whatever the
size of the problem which is very promising.  Moreover, for magic
squares the speedup tends to increase as the problem becomes more
difficult which is also a very interesting property.

This evaluation has also uncovered an even more significant improvment
on the worst case: the obtained speedup is always better than the one
obtained in the average case.  Like this, \PROG{} greatly narrows the
range of possible execution times for a given problem.
Figure~\ref{all-interval-450-execs-graph} depicts the graph of the 50
executions for the all-interval 450 benchmark, both with 1 and 16 SPUs
(due to space limitation we only show this one but a similar graph
exists for all other problems).  This graph clearly reveals the
difference of dispersion when using 1 SPU or 16 SPUs.
Table~\ref{stdev} charts the evolution of the standard deviation of
the execution times for the largest instance of each problem depending
on the number of SPUs.  The standard deviation rapidly decreases when
more SPUs are used (the most spectacular case being magic square $100
\times 100$ where the standard deviation decreases from 915.7 to 3.8).
\PROG{} limits the dispersion of the execution times.  We can say that
the multicore version is more robust than the sequential one in the
sense that the difference between the minimum and maximum execution
times, as well as the overall variance of the results, decreases
significantly.  Therefore, the execution time is more predictable from
one run to another in the multicore version, and more cores means more
robustness. This is crucial for real-time systems or even some
interactive applications.


\begin{table}[htbp]
\begin{center}
\begin{tabular}{||c|r|r|r|r||}
\hline
     & \C{all}      & \C{number} & \C{perfect} & \D{magic} \\
SPUs & \C{interval} & \C{partit} & \C{square}  & \D{square} \\
     & \C{450}      & \C{2600}   & \C{5}       & \D{100} \\
\hline\hline
 1   & 891.2 & 24.0 & 122.0 & 915.7  \\
 2   & 459.6 & 16.0 &  54.4 &  18.5  \\
 4   & 223.9 &  6.9 &  26.6 &  12.7  \\
 8   &  65.4 &  2.8 &  16.3 &  10.0  \\
12   &  61.0 &  1.9 &  10.4 &   3.4  \\
16   &  40.0 &  1.5 &   6.3 &   3.8  \\
\hline
\end{tabular}
\end{center}
\caption{evolution of the standard deviation (50 execution times)}\label{stdev}
\end{table}

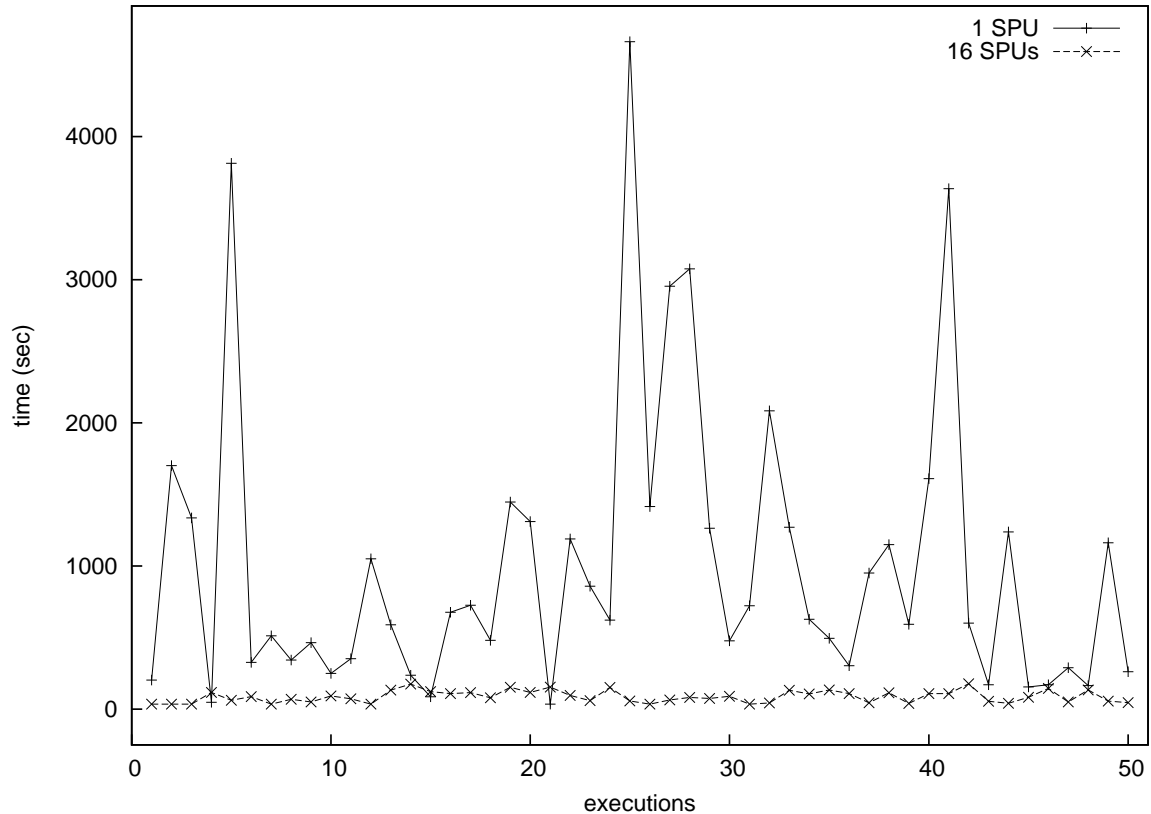
\AddGraph{dispersion analysis for the 50 runs of all-interval 450}{all-interval-450-execs}{1}

An interesting experiment can be made to further develop this idea.
We experimented with a slight variation of the method which consists
in starting all parallel processes with the \emph{same} initial
configuration (instead of a random one).  Each SPU will then diverge
according to its own internal random choices.  We implemented this
variant and the results show that the overall behavior is practically
the same as the original method, just a bit slower on average by about
10\%.  This slowdown was to be expected because in this case the
search has less diversity to start with, and therefore might take
longer to explore a portion of the search space that contains a
solution.  However the fact that this slowdown is only 10\% shows that
the method is intrinsically quite robust, can restore diversification
and take again advantage of the parallel search in a quite efficient
manner.

Finally, it is worth noticing that Adaptive Search is an ``anytime'' method:
it is always possible to interrupt the execution and to obtain the best
pseudo-solution computed so far. On this point too this method can benefit
easily from the Cell: when running several SPUs in parallel, the PPE simply
has to ask each SPU to obtain its best pseudo-solution (together with the
corresponding cost) and then to chose the best of these bests. Indeed,
another good property regarding the Cell features, is the fact that the only
data a SPU needs to pass is the current configuration (an array of integers)
and the associated cost.

\section{Concluding Remarks}
\label{sec:conclusion}

We presented a simple yet effective initial port of the Adaptive
Search algorithm to the Cell/BE architecture, which we used to solve
combinatorial search problems.  The experimental evaluation we carried
out indicates that linear speedups are to be expected in most cases,
and even some situations of superlinear speedups are possible.
Scaling the problem size seems never to degrade the speedups, even
when dealing with very difficult problems.  We even ran a reputedly
very hard benchmark with increasing speedups when the problem size
grows.

An important, if somewhat unexpected, fringe benefit is that the worst
case execution time gets even higher speedups than the average case.
This characteristic opens up several domains of application to the use
of combinatorial search problem formulations: this is particularly
true of real-time applications and other time-sensitive usages, for
instance interactive games.

Clearly the Cell/BE has a very significant potential to make good on
combinatorial search problems.  We plan to work on two separate
directions: on one hand, to optimize the code as per the IBM
guidelines~\cite{cell} and on the other, to experiment with more
sophisticated organizations and forms of communication among the
processors involved in a computation.

\section*{Acknowledgements}

The equipment used to perform the benchmarks described herein was
provided by IBM Corporation, as a grant from the Shared University
Research (SUR) program awarded to CENTRIA and U.~of \'Evora.

\bibliographystyle{eptcs}
\bibliography{lscs09-final}

\end{document}